\documentclass[a4paper, 10pt, conference]{ieeeconf}   
\pdfoutput=1   
\usepackage{FG2021}
\usepackage{subfiles}
\usepackage{graphicx}
\usepackage{comment}
\usepackage{amsmath,amssymb} 
\usepackage{color}
\usepackage{subfiles}
\usepackage{subfig} 
\usepackage{booktabs}
\usepackage{url}
\usepackage{caption}
\usepackage{lipsum}
\newcommand{\etal}{\textit{et al. }}
\FGfinalcopy

\IEEEoverridecommandlockouts                              
\def\FGPaperID{119} 

\title{\LARGE \bf
Temporally coherent video anonymization through GAN inpainting
}

\author{\parbox{16cm}{\centering
    {\large Thangapavithraa Balaji$^1$, Patrick Blies$^2$, Georg G\"ori$^2$, Raphael Mitsch$^2$,
    Marcel Wasserer$^2$ and Torsten Sch\"on$^3$}\\
    {\normalsize
    $^1$Audi AG, thangapavithraa.balaji@audi.de, $^2$Enlite.ai\\
    $^3$Technische Hochschule Ingolstadt, torsten.schoen@thi.de}}
}

\begin{document}

\ifFGfinal
\thispagestyle{empty}
\pagestyle{empty}
\else
\author{Anonymous FG2021 submission\\ Paper ID \FGPaperID \\}
\pagestyle{plain}
\fi
\maketitle

\begin{abstract}

This work tackles the problem of temporally coherent face anonymization in natural video streams.
We propose \emph{JaGAN}, a two-stage system starting with detecting and masking out faces
with black image patches in all individual frames of the video.
The second stage leverages a privacy-preserving Video Generative Adversarial Network
designed to inpaint the missing image patches with artificially generated faces.
Our initial experiments reveal that image based generative models are not capable of inpainting
patches showing temporal coherent appearance across neighboring video frames.
To address this issue we introduce a 
newly curated video collection, which is
made publicly available for the research community along with this paper\footnote{Download script is available at: \url{https://github.com/cvims/jagan}}.
We also introduce the Identity Invariance Score IdI as a means to 
quantify temporal coherency between neighboring frames.

\end{abstract}

\begin{figure*}[tb]
  \centering
    \includegraphics[width=0.75\textwidth]{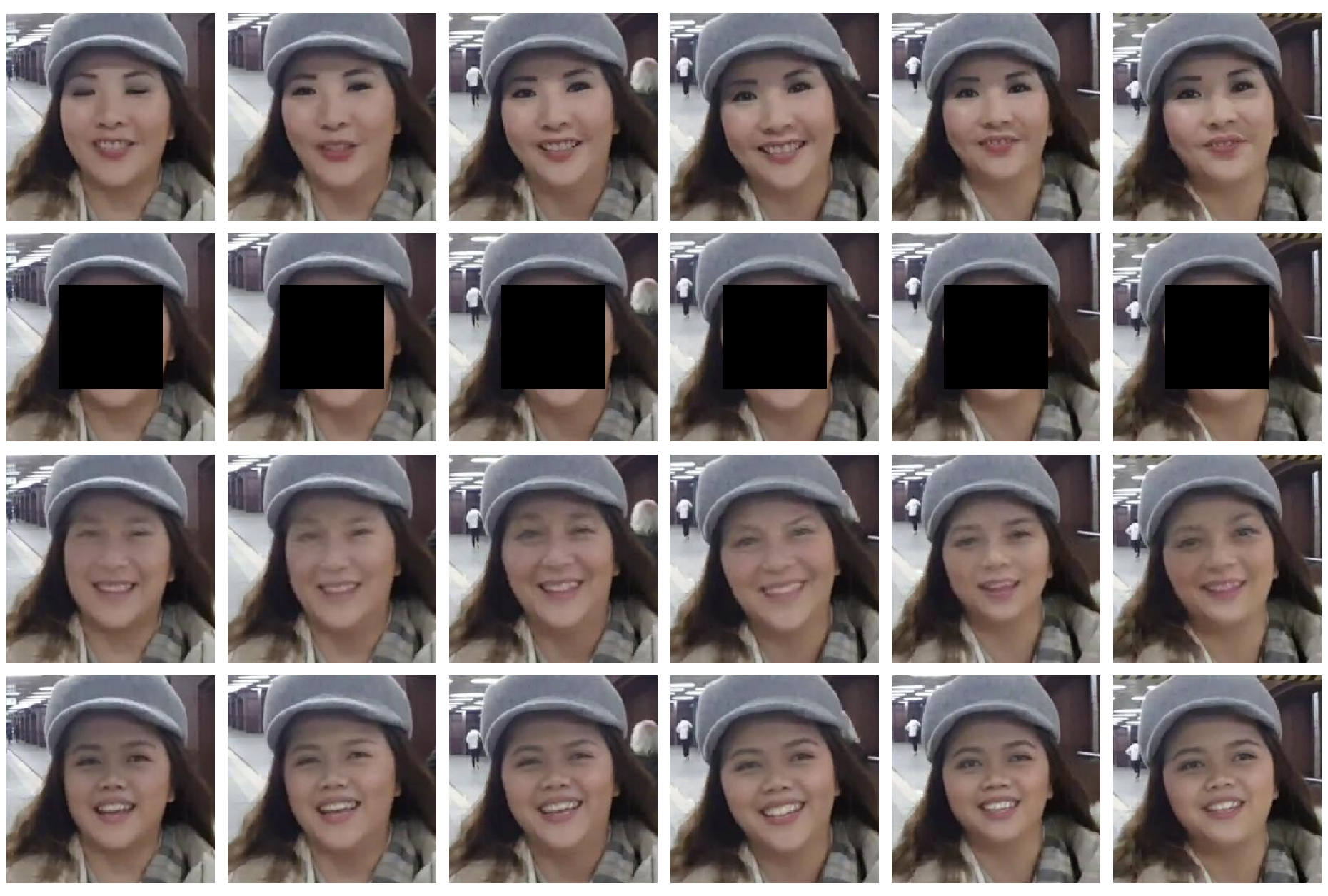}
  \caption{Video dataset and generation examples (skipping 5 frames between each column).
  First row: real images. Second row: anonymization areas and conditional input for our generator.
  Third row: images generated with an image--based model.
  Note the shift in the identity of the 
  generated face. Fourth row: video generative model enforcing temporal coherence. The 
  identity of the generated faces remains the same across frames.}
  \label{fig:example_videos}
\end{figure*}

\section{Introduction}
\label{sec:intro}
Autonomous systems and applications relying on deep learning based image processing
require to be trained on data which is collected in public areas such as rural street scenes. One prominent example in this context is the development of autonomous vehicles (e.g. self driving cars). Naturally, this data can include persons which did not explicitly accept the collection of their data or even might not be aware of it.
To protect people's personal rights, the European Union adopted the General Data Protection Regulation (GDPR) in 2016 \cite{eu-32016R0679}. As a consequence of this regulation, it is required to anonymize persons in images taken from public areas which is typically done by cutting out, blurring or pixelating human faces \cite{Neustaedter2003BalancingPA}\cite{10.1145/330534.330540}\cite{10.1145/240080.240295}. Although anonymization is ensured, these approaches come with the major drawback of modifying the original image structure. When dealing with temporal video data, this effect is even more visible, as individually anonymized frames leave a flickering effect on the video sequence. In this paper, we introduce a temporally coherent video anonymization technique to solve these issues called JaGAN, the faceless network.     

\subsubsection{Our Contributions}

Our proposed method follows a two step approach
by first identifying and masking out faces in individual frames
and then inpainting the missing content with an artificially generated new face using a temporally coherent Generative Adversarial Network \cite{NIPS2014_5423}.
We demonstrate that using such a method not only allows to generate consistent faces across several image frames within a video sequence, but also leads to more natural looking faces and a higher FID score \cite{DBLP:conf/nips/HeuselRUNH17} even in single images compared to state of the art methods
\cite{ganonymizer}\cite{DBLP:conf/isvc/HukkelasML19}\cite{stngan}.
To ensure a full anonymization without leaking any information from the original face, our approach avoids using any facial characteristics or features such as landmarks. Results in Section~\ref{image_models} demonstrate that with our proposed method we improve the state of the art in landmark--free facial inpainting. 
In order to measure and compare the temporal coherency of generated faces along a sequence of frames, we introduce an identity invariance score in Section~\ref{identity_invariance}.
Our results on single frames are evaluated on the FDF dataset \cite{DBLP:conf/isvc/HukkelasML19} whereas for evaluating the temporal coherence video sequences of faces are required.

For this purpose, we present and release a novel large scale video dataset comprising approximately 400.000 sequences with a length of 30 frames each.
All sequences in the collection have been crawled from Youtube
-- considering only videos released under the Creative Commons Licence --
to provide a publicly available dataset for training and evaluation of temporally coherent generative inpainting methods.
Fig.~\ref{fig:example_videos} shows an example of a video contained in the dataset along with the inpainting results of our image-based as well as video-based anonymization GAN.

\section{Related Work}
\label{sec:related_work}
Ad hoc obfuscation techniques like blurring, pixelation and various other filters \cite{Neustaedter2003BalancingPA}\cite{10.1145/330534.330540}\cite{10.1145/240080.240295} are widely used to anonymize privacy sensitive information of individuals in single images.
However, images anonymized using these techniques are susceptible to loss of quality and utility.
McPherson \etal \cite{DBLP:journals/corr/McPhersonSS16} demonstrate that persons anonymized using these techniques can also be identified using face recognition algorithms due to their robustness.
Many ad hoc methods, to a certain degree, fail to remove privacy sensitive information and prove to be inadequate in de-identifying the persons in the images  \cite{Ribaric2015AnOO}\cite{10.1145/358916.358935}\cite{10.1109/TKDE.2005.32}. 

Since the introduction of Generative Adversarial Nets \cite{NIPS2014_5423}, there has been several attempts to adapt GAN models for tasks such as text-to-image generation \cite{8237891}\cite{DBLP:journals/corr/ReedAYLSL16}, domain transfer \cite{CycleGAN2017}\cite{pix2pix2016}\cite{8100223}, super resolution image generation \cite{DBLP:journals/corr/abs-1710-10196}\cite{8099502} and image completion purposes \cite{DBLP:journals/corr/MirzaO14}\cite{DBLP:conf/cvpr/LiLY017}\cite{DBLP:conf/cvpr/Yeh0LSHD17}.

The scope of this paper is to reuse the central idea of GANs to design a model for the task of temporally coherent video anonymization.

\paragraph{Face Completion}
Face completion is a more challenging task compared to image completion, since for image completion the model can pick up contextually similar patterns to reuse them for filling up the missing regions.
In contrast, face completion tasks require the generation of unique patterns and details not present in the input image (e.g., generating lips, eyes) \cite{DBLP:conf/cvpr/LiLY017}\cite{DBLP:conf/cvpr/Yeh0LSHD17}.
The drawback of the works above is that the results are not convincing in terms of quality, anonymization (which was not their scope) and especially real-world applicability as the dataset used is CelebA containing only frontal facing faces.
Wu \etal \cite{stngan} propose an approach to perform inpainting with spatio-temporal consistency in videos.

\paragraph{Anonymization with Face Generation}
 Ren \etal \cite{DBLP:conf/eccv/RenLR18} propose a video anonymization method to modify the privacy sensitive data of individuals on pixel level.
 Shirai \etal \cite{DBLP:conf/cvpr/ShiraiW19} demonstrate the concept of generating a surrogate image and performing a conditional based training.
 Wang  \etal \cite{DBLP:conf/nips/vid2vid} propose the concept of video-to-video synthesis with conditional style transfer.
 The spatio-temporal adversarial objective introduced enforces the generation of temporally coherent videos.
 CIAGAN \cite{DBLP:conf/cvpr/MaximovEL20} not only performs face inpainting but 
 controls the de-identification procedure as well. 
 DeepPrivacy \cite{DBLP:conf/isvc/HukkelasML19} is most similar to our work and serves as a basis for the development of our method as well as our experimental evaluation.
 The conditional generator allows to maintain a seamless transition between the generated face and the surrounding background.
 
 \begin{figure*}[t!]
  \centering
    \includegraphics[width=0.96\textwidth]{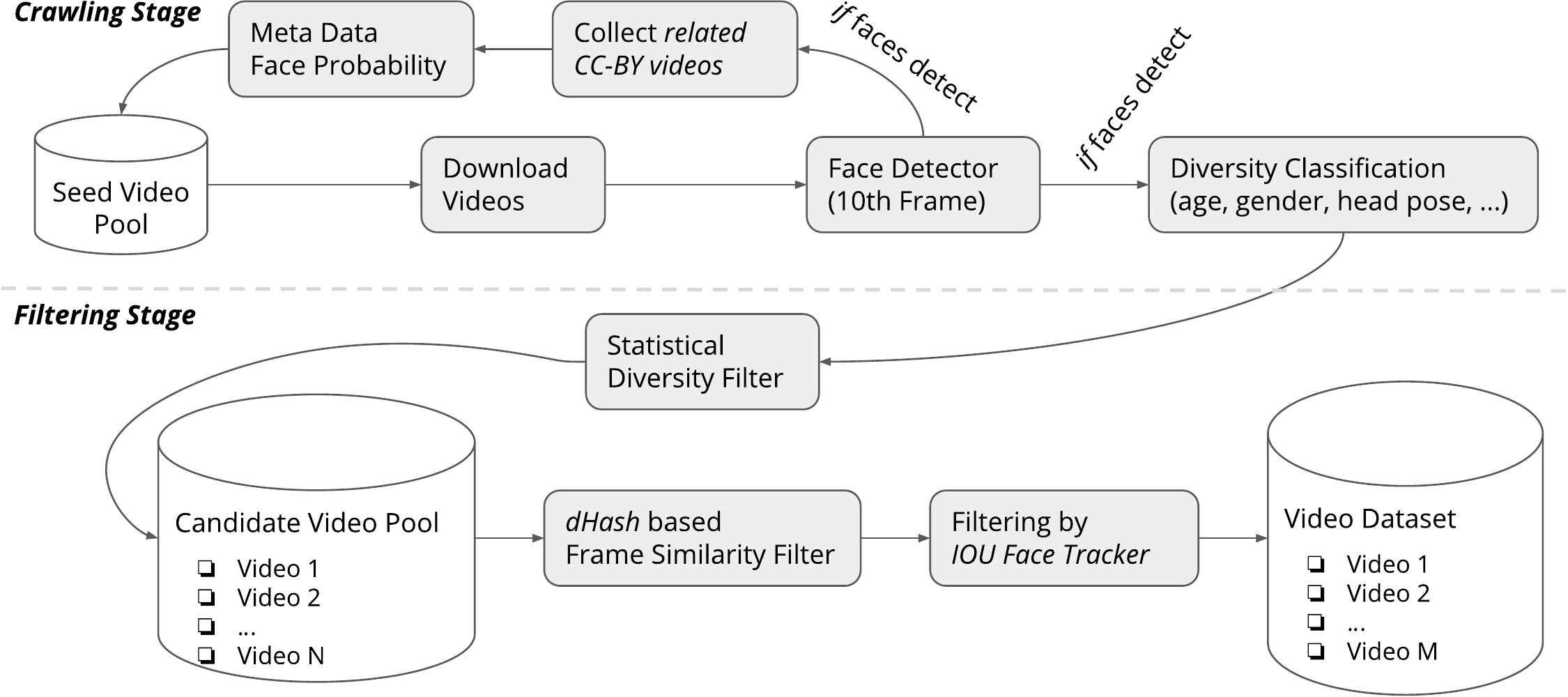}
  \caption{Overview of video data set collection pipeline.}
  \label{fig:data_set_collection}
\end{figure*}

\section{Datasets}
\label{sec:video dataset}
This section describe the image and video data used in our evaluations.

\subsection{FDF Image Dataset}
The Flickr Diverse Dataset (FDF) has been crawled from the website Flickr and contains 1.47 Million faces in the wild with a minimum resolution of 128\(\times\)128 pixels. The faces have been automatically annotated with 7 facial landmarks as well as with a bounding box and have a large diversity in terms of age, ethnicity, facial painting, facial pose, occluding objects and image background. FDF is implicitly designed for training generative models with the purpose of face anonymization. Further details are given by Hukkelas \etal{}\cite{DBLP:conf/isvc/HukkelasML19}.

\subsection{Video Dataset}
We curate and release a large scale video dataset of human faces in 99,795 different videos to enable the training and evaluation of
video based person anonymization methods as proposed in Section~\ref{sec:video_method}. The Creative Commons licensed videos (\emph{CC-BY}) are crawled from public YouTube
channels and selected to preserve diversity in terms of gender, age, ethnicity and poses. The data collection
was performed in a 2-stage process:
first, we collected a dataset of \emph{seed faces},
for which we then collected videos at a later stage.
Fig.~\ref{fig:data_set_collection} shows and overview of the entire data set collection pipeline
for which we describe the individual components in detail below.

\subsubsection{Seed Faces Collection Process (Crawling Stage)}
In this section we describe our pipeline to collect \emph{relevant} content from YouTube.
By relevant, we refer to a training data set comprising images of faces
that show a sufficiently large variability with respect to:

\begin{itemize}
  \item faces with a large variety of angles and pitches
  \item faces with a large variety of ages
  \item female and male faces
  \item faces from people of different ethnicity
\end{itemize}

To start, we manually specify a set of seed videos
for which we expect to have relevant facial examples
for the anonymization task.
For the first set of seed videos, we mainly used dashcam and outdoor videos.
While running the pipeline, the list of seed videos is
progressively extend by videos that have detected faces in them.
This way, we end up having theoretically infinite supply of seed videos.
At a later stage, after evaluation of the feature class distributions,
we manually add queries to the seed video pool to obtain more instances of 
underrepresented categories.

For each video in the seed pool,
we start by extracting the corresponding metadata
and feed it as an input to a binary classifier
predicting a probability that the video actually contains faces.
This metadata pre-classification stage is video (image) agnostic and allows for an efficient way to filter the set of candidate videos.
All videos passing this filter are forwarded to the subsequent stage of actual video analysis.
Note, that the metadata classifier is repeatedly retrained with the
growing set of actually analyzed videos serving as ground truth training data.

\begin{figure*}[t!]
  \centering
    \includegraphics[width=0.98\textwidth]{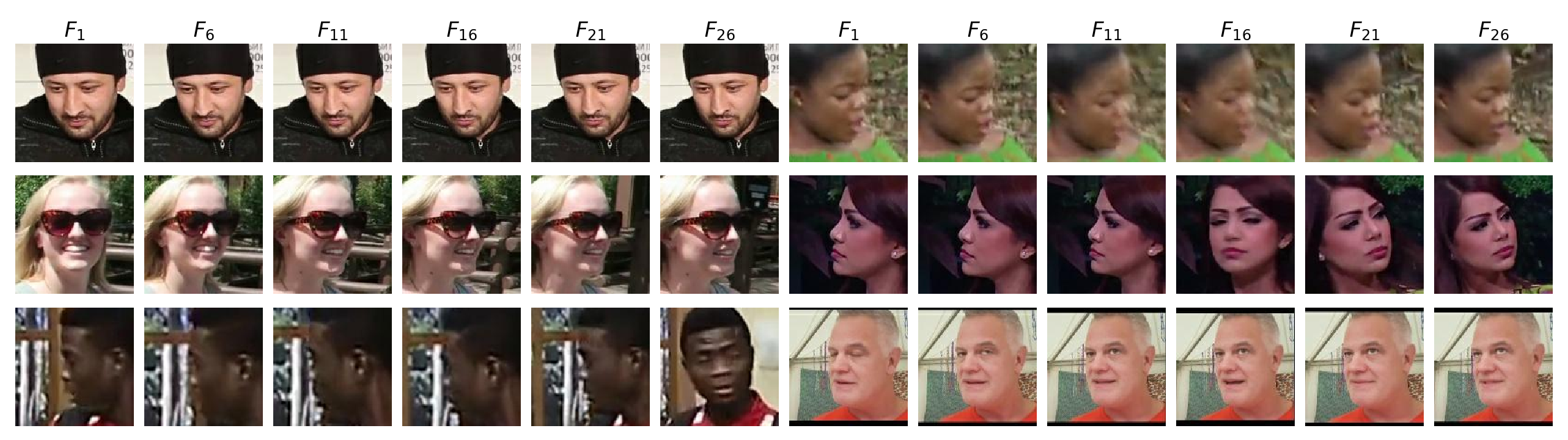}
  \caption{Example videos contained in the dataset. (For space reasons we only show every fifth time step of each 30 frame sequence.)}
  \label{fig:example_videos}
\end{figure*}

Each video passing the previous stage is further analysed,
by extracting the 10th frame (empirically chosen) and by running a
face detector (we have used a modified version of TinyFaces face detector \cite{Hu_2017_CVPR}). 
If at least one face is found in this frame,
we search YouTube for \emph{related videos} and add those to the pool of seed videos.
The faces detected in these videos are collected and
added to a collection of seed faces.

Given the collection of seed face images
we utilize a set of simple image classifiers tagging the individual images
with the following classes for Ethnicity (\emph{asian, black, white, other}), Age (\emph{0-16, 16-50, 50+}) and Gender (\emph{female, male}).
For developing these image categorization models we use the UTKFace Dataset\cite{DBLP:conf/cvpr/ZhangSQ17}. 
Additional we estimate the head rotation using PRN\cite{DBLP:conf/eccv/FengWSWZ18} and put the results into the following buckets (\emph{front, half-pose, portrait})
\footnote{We are aware that this automatic categorization is not ideal but keeping the scale of our dataset in mind so far the only feasible option.}.

For training, validation and test set, we select an equally distributed subset 
of sequences to contribute to the final dataset of seed faces. To reduce the risk of leakage of identities across the dataset splits, we used a single YouTube channel only for one split.

\subsubsection{Video Collection Process (Filtering Stage)}
Given the pool of seed faces the video candidates are further analysed by trying to track these faces in the subsequent frames
with an IOU Tracker \cite{DBLP:conf/avss/BochinskiES17}.
If tracking is possible for at least 30 frames
we further filter the sequences by ensuring
that each frame in a series has a high \emph{dHash} \cite{dHash} similarity score.
This last step discards sequences comprising abrupt changes such as a scene cuts.
Regarding the minimum count of 30 frames we follow \cite{DBLP:conf/nips/vid2vid}
serving as a basis for our video based anonymization models.

When compiling the dataset we crop centered on the face bounding box and scaled to size of 256px.
The total number of faces contained in the final dataset to be released is shown in Table~\ref{tab:table2},

Fig.~\ref{fig:example_videos} shows a representative subset of the frames contained in the dataset.

\begin{table}[tb]
\caption{Size of train, validation and test sets}
\centering
\begin{tabular}{lcc}
\toprule
Set & No. of initial frames & Total no. of faces \\
\midrule
Train & 348,987 & 10,469,610\\
Validation &32,268 & 968,040\\
Test &33,571 & 1,007,130\\
\bottomrule
\end{tabular}
\label{tab:table2}
\end{table}

\section{GAN Based Image Anonymization}
\label{sec:image_method}
This section describes the architecture of our image based anonymization GAN
along with the required pre--processing steps as well as the applied training procedure.
Note that the scope of this work is not to improve the state of the art in image anonymization
but to introduce temporal coherence when working with video data.
We introduce and compare our model with a reference method \cite{DBLP:conf/isvc/HukkelasML19}
at this point to make sure that our design choices serve as a valid basis
for the video anonymization network proposed in Section~\ref{sec:video_method}.  

\subsection{Architecture}
In general, our architecture (see Fig.~\ref{fig:image_model}) is an adaptation of \cite{DBLP:conf/cvpr/Wang0ZTKC18} combined with ideas from other works such as the 
2-stage approach proposed in \cite{DBLP:conf/cvpr/Yu0YSLH18}, the 
spatially discounted reconstruction loss from \cite{DBLP:conf/cvpr/Yu0YSLH18}
and $R_1$ regularization from \cite{DBLP:conf/icml/MeschederGN18}.

When training on the FDF dataset, input images have the size $128 \times 128$,
whereas for the proposed video dataset, 
the input images have the size $256 \times 256$.
In the following, we denote the images dimensions simply with size $h \times w$.

\subsubsection{The Coarse Stage}
\label{sec:coarse-stage}
\begin{figure}[tb]
    \includegraphics[width=\linewidth]{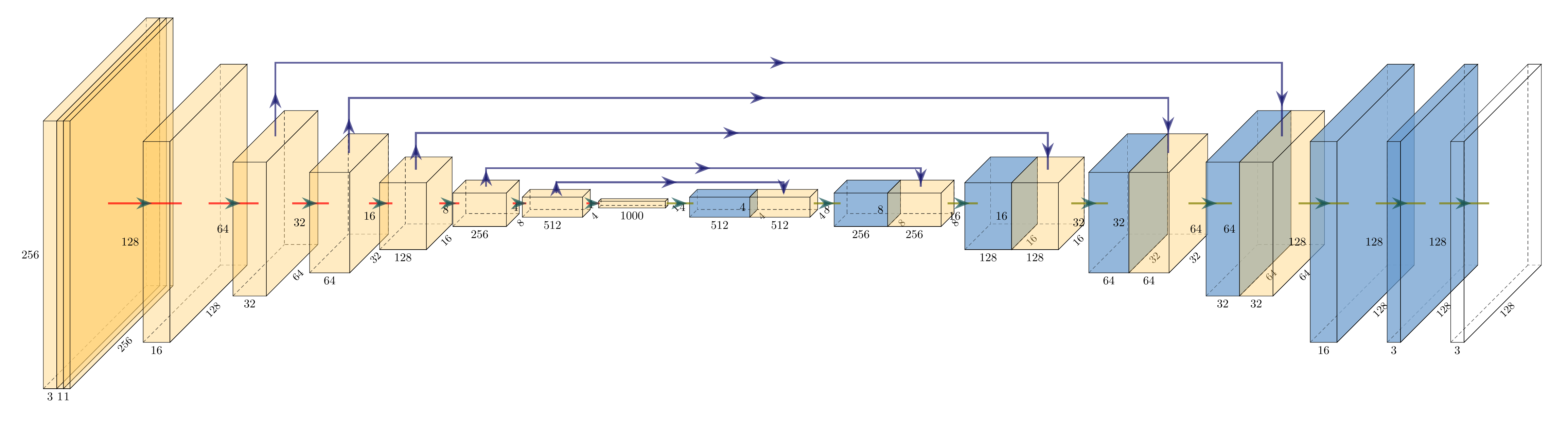}
    \caption{Depiction of our coarse stage U--Net architecture for $256 \times 256$ input images. Yellow boxes
      correspond to feature maps after convolutions, blue boxes correspond to
      feature maps after transpose
      convolutions. The transparent box corresponds to a feature map after a
      tanh layer. Green arrows correspond to $4\times4$ (transpose) convolutions
      with batch normalization and Leaky ReLUs.
    }\label{fig:coarse}
\end{figure}

The coarse stage is based
on the U--Net architecture \cite{DBLP:conf/miccai/RonnebergerFB15}
as shown in Fig.~\ref{fig:coarse}.
From the pre--processing stage, we obtain a tensor of shape $(5, h, w)$\footnote{We omit the batch size dimension for simplicity.}.
This corresponds to an image with three color channels, one channel for the border
mask and one channel for the anonymization area mask.
As depicted in Fig.~\ref{fig:coarse},
several strided convolution--batch norm--leaky
ReLU operations are applied to the input tensor.
Each red arrow in the figure corresponds to a $4 \times 4$ convolution with stride 2 followed by batch normalization and a leaky ReLU activation function.
The bottleneck layer has dimensions $(1000, 1, 1)$.
After the bottleneck layer, we feed the tensor through several
transposed convolution--batch norm--ReLU-concat skip connections operations.
Each yellow arrow corresponds to a $4\times4$ transposed convolution with batch normalization and ReLU activation function. Each blue error
corresponds to a copy--and--concat operation.
The last layer in the coarse stage is a $\tanh$--layer.

\subsubsection{The Fine Stage}
\label{sec:fine-stage}
\begin{figure}[tb]
    \includegraphics[width=\linewidth]{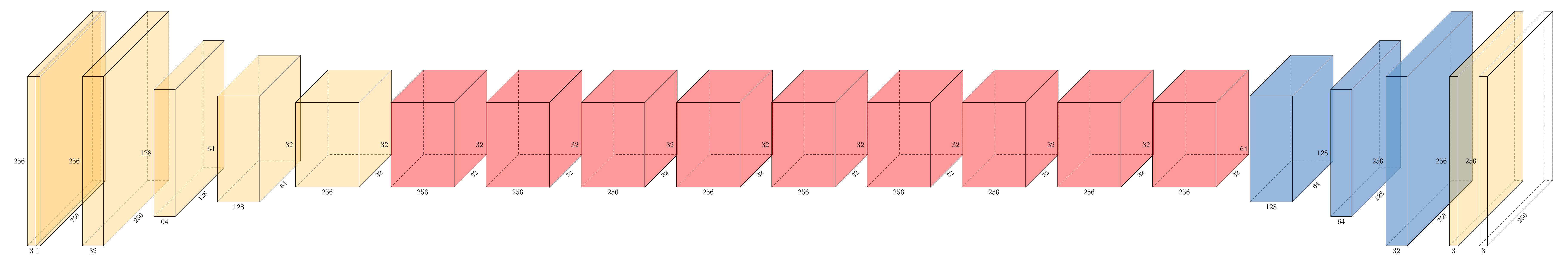}
    \caption{Depiction of our fine stage architecture. Yellow boxes
      correspond to feature maps after convolutions, red boxes correspond to
      feature maps after 
      residual blocks, blue boxes correspond to
      feature maps after transpose
      convolutions. The transparent box corresponds to the feature map after a
      tanh layer.}
    \label{fig:image_model}
\end{figure}

The fine--stage generator network consists of three components:
a convolutional front-end with strided convolutions (yellow in Fig.~\ref{fig:image_model}),
a set of nine residual blocks (red in Fig.~\ref{fig:image_model}),
a transposed convolutional back--end (implemented as upsampling + convolution,
blue in Fig.~\ref{fig:image_model}) and finally a tanh output layer.
The input to the fine stage again is a tensor of shape $(5, h, w)$.
As the output of the coarse stage is only of size $(5, h/2, w/2)$ as proposed in \cite{DBLP:conf/cvpr/Yu0YSLH18} we rescale it to original input resolution $(5, h, w)$. Afterward we again combine it with the border and anonymization area mask before feeding it into the fine stage network.

\subsubsection{The Discriminator}
For the discriminator, we use the multi-scale discriminator approach of \cite{DBLP:conf/cvpr/Wang0ZTKC18}.
Like them, we use three discriminators with identical network structure
operating at different image scales, leading to an increased receptive field of the discriminator.
The real and synthesized
images ($h \times w$ in our case) are down sampled by a factor
of two and four, yielding images with resolutions $h/2 \times h/2$ and $h/4\times h/4$ respectively.
Each of the three discriminators is then trained to differentiate
real and fake images at one of these  scales.
This way, the discriminator operating on the $h / 4\times h/4$ images
has the largest receptive field and can guide the generator to generate
globally consistent images while the discriminator operating on $h \times w$ 
encourages the discriminator to produce finer details.

\begin{figure}[b]
\subfloat[]
{
  \begin{minipage}[h]{0.17\linewidth}
    \centering
    \includegraphics[width=0.99\linewidth]{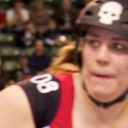}
     \label{fig:pipeaa}
  \end{minipage}
  }
\subfloat[]
{
  \begin{minipage}[h]{0.17\linewidth}
    \centering
    \includegraphics[width=0.99\linewidth]{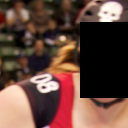}
     \label{fig:pipebb}
  \end{minipage}
  }
\subfloat[]
{
  \begin{minipage}[h]{0.20\linewidth}
    \centering
    \includegraphics[width=0.99\linewidth]{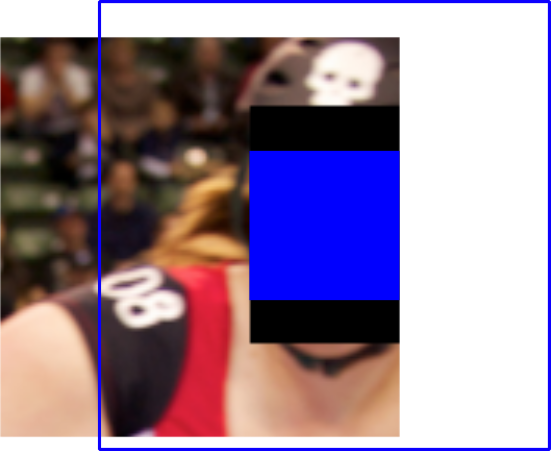}
     \label{fig:pipecc}
  \end{minipage}
}
\subfloat[]
{
  \begin{minipage}[h]{0.17\linewidth}
    \centering
    \includegraphics[width=0.99\linewidth]{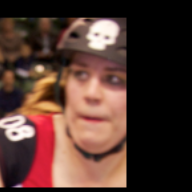}
     \label{fig:pipedd}
  \end{minipage}
}
\subfloat[]
{
  \begin{minipage}[h]{0.17\linewidth}
    \centering
    \includegraphics[width=0.99\linewidth]{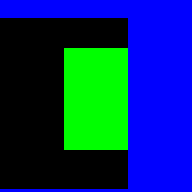} 
    \label{fig:border_mask}
  \end{minipage}
}

\caption{Pre--processing steps for our image model}
  \label{fig:pipeline}
\end{figure}

\subsubsection{Pre--Processing}
\label{sec:preproc_images}
The pre--processing pipeline is summarized in Fig.~\ref{fig:pipeline}. 
First, faces are detected with DSFD.
This yields bounding boxes for all faces in an image.
Everything within the detected box is deleted (replaced with black pixels),
resulting in the image shown in Fig.~\ref{fig:pipebb}.
To get the context rectangle, the 
area of the bounding box is made quadratic by reducing the length of the larger edge (Fig.~\ref{fig:pipebb}) followed tripling the 
edge length of the resulting square (blue outline in Fig.~\ref{fig:pipecc}) 
Everything within the resulting square 
area is taken as context for the generator. 
If the enlarged bounding box extents beyond the border of the image,
we add a corresponding region to the original image (Fig.~\ref{fig:pipedd}).
The resulting image is centered on the face.
We provide the generator with additional information about the image border
by supplying it with a border mask (blue in Fig.~\ref{fig:border_mask}).
To indicate which area needs to be inpainted,
we introduce an additional anonymization mask 
(green in Fig.~\ref{fig:border_mask}). 
The masks are concatenated to the image along the channel dimension,
yielding our input tensor for the generator.
In contrast to the model presented in \cite{DBLP:conf/isvc/HukkelasML19}
we avoid feeding facial landmarks as an input into the generator for two reasons:
First, even though the image patch is replaced with black pixels,
the position of landmarks might still leak information about the identity of the anonymized individual.
Second, when extending the model towards video anonymization in Section~\ref{sec:video_method},
inconsistent and noisy landmark detections in subsequent frames
have a direct impact on the respective, generated video frames.
In particular, false landmark detections lead to temporally inconsistent video frames
with respect to facial features of the generated faces.
\subsubsection{Optimization Details}
\paragraph{Loss Functions}
After experimenting with different adversarial loss functions,
such as the vanilla GAN loss \cite{DBLP:conf/nips/GoodfellowPMXWOCB14},
WGAN \cite{DBLP:journals/corr/ArjovskyCB17}
and hinge--loss \cite{DBLP:journals/corr/TranRB17},
we finally decided to apply the least squares loss from LSGAN
\cite{DBLP:conf/iccv/MaoLXLWS17}
producing the most convincing results in terms of qualitative performance.

Another idea we reuse from \cite{DBLP:conf/cvpr/Wang0ZTKC18} is that of the feature matching loss.
The idea behind that loss term is to use the output of
each layer of each of the three multi-scale discriminators to calculate the mean $L_1$-distance
between the activation of generated and real images.
This addition term helps to stabilize training by encouraging the generator
to produce natural statistics at multiple scales for all layers of the discriminator.

In summary, we use the following loss components: For the coarse stage reconstruction loss and for the fine stage adversarial LSGAN loss, feature matching loss and spatially discounted reconstruction loss, additionally we regularize the discriminator with $R_1$  \cite{DBLP:conf/icml/MeschederGN18}.

\paragraph{Training Procedure}
For training we apply the two time--scale update rule from \cite{DBLP:conf/nips/HeuselRUNH17} and optimize the networks using Adam \cite{DBLP:journals/corr/KingmaB14}
with a learning rate of 0.0004 and a batch size of 256.
We track the FID score \cite{DBLP:conf/nips/HeuselRUNH17} on the validation set
for early stopping and best model selection. The selected model was trained over 89,928 mini-batches.

\section{Temporally Coherent Video Anonymization}
\label{sec:video_method}
This section proposes a method for temporally coherent video anonymization. 
Starting from the image based model described above we establish temporally consistent generation results by
implementing the temporal consistency terms added
by \cite{DBLP:conf/nips/vid2vid}. Additional we apply \emph{burn-in}, a technique to improve inference results for the starting frames of a video.

\subsection{Architecture}
The fine stage and the training regime described in Section \ref{sec:image_method} are reused for the video model.
However, in the discriminator and the input data representation we introduce major changes to support temporal coherence across the generated video frames. In addition to the existing image discriminator we use an additional video discriminator as described below.

\subsubsection{Conditional Video Discriminator}
Vid2Vid \cite{DBLP:conf/nips/vid2vid} proposes several methods to enforce temporal 
consistency, one of them being the conditional video discriminator,
which we also employ for our video model.
In summary, its purpose is to ensure that consecutive frames resemble the temporal dynamics
of a real video, thereby enforcing temporal consistency between the frames.
It does so by distinguishing between three frames of a generated and a real sequence.
The video discriminator is
multi--scale in space and time: the frame rates of the real and generated videos
are downsampled by a factor of 3, skipping in--between frames in the process.
This is done for up to three time scales ($(t_0,t_{-1},t_{-2})$, $(t_0,t_{-3},t_{-6})$) and $(t_0,t_{-9},t_{-18})$) depending on how many generated frames are already available.
The different time-scales encourage short-- and long--term consistency. Together with 3 spatial scales per time scale, this results in up to 9 discriminators, which are used in training.

\subsection{Pre--processing}
Since we have conditionally strongly dependent frames in each sequence, one step to enforce temporal consistency is 
to extend the input that is fed into the generator network by adding past frames to the input tensor. This idea was proposed 
in \cite{DBLP:conf/nips/vid2vid}.
They use two images from the past to condition the generator on the previous sequence.
The reason for using two images is that they claim that fewer than two images lead to unstable results, while more than two images lead to a higher GPU workload and memory footprints without increasing quality. With this addition, the pre--processing 
pipeline changes in the following way: instead of feeding one image (+ anonymization and boundary masks) to the generator, we now feed three images (+ anonymization and boundary masks) to the generator: two images from the past and the current image that is to be inpainted at this step.

\subsection{Burn-in Stage}
\begin{figure}[tb]
\subfloat[]
{
  \begin{minipage}[h]{0.32\linewidth}
    \centering
    \includegraphics[width=0.9\linewidth]{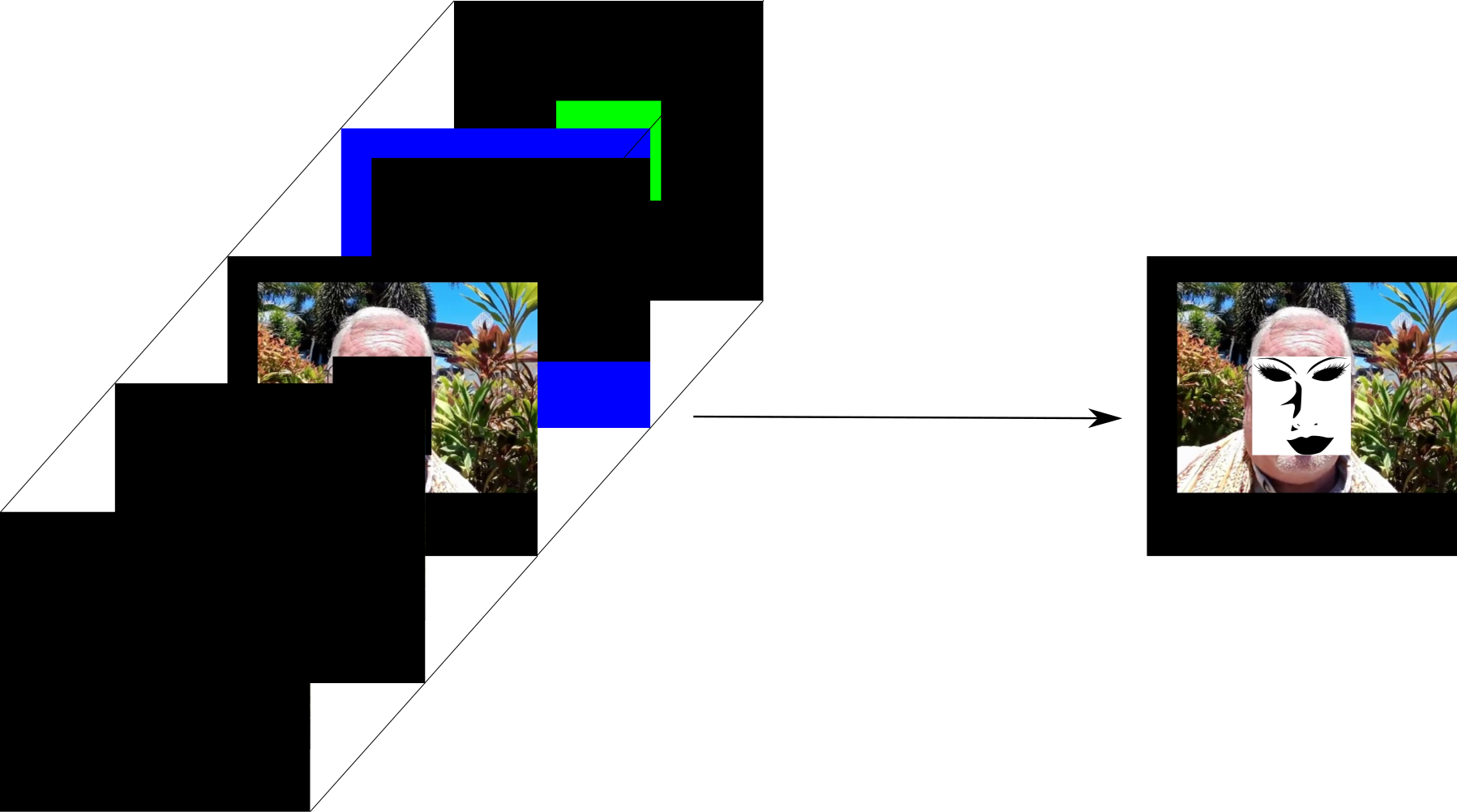}
     \label{fig:burn_in_a}
  \end{minipage}
  }
\subfloat[]
{
  \begin{minipage}[h]{0.32\linewidth}
    \centering
    \includegraphics[width=0.9\linewidth]{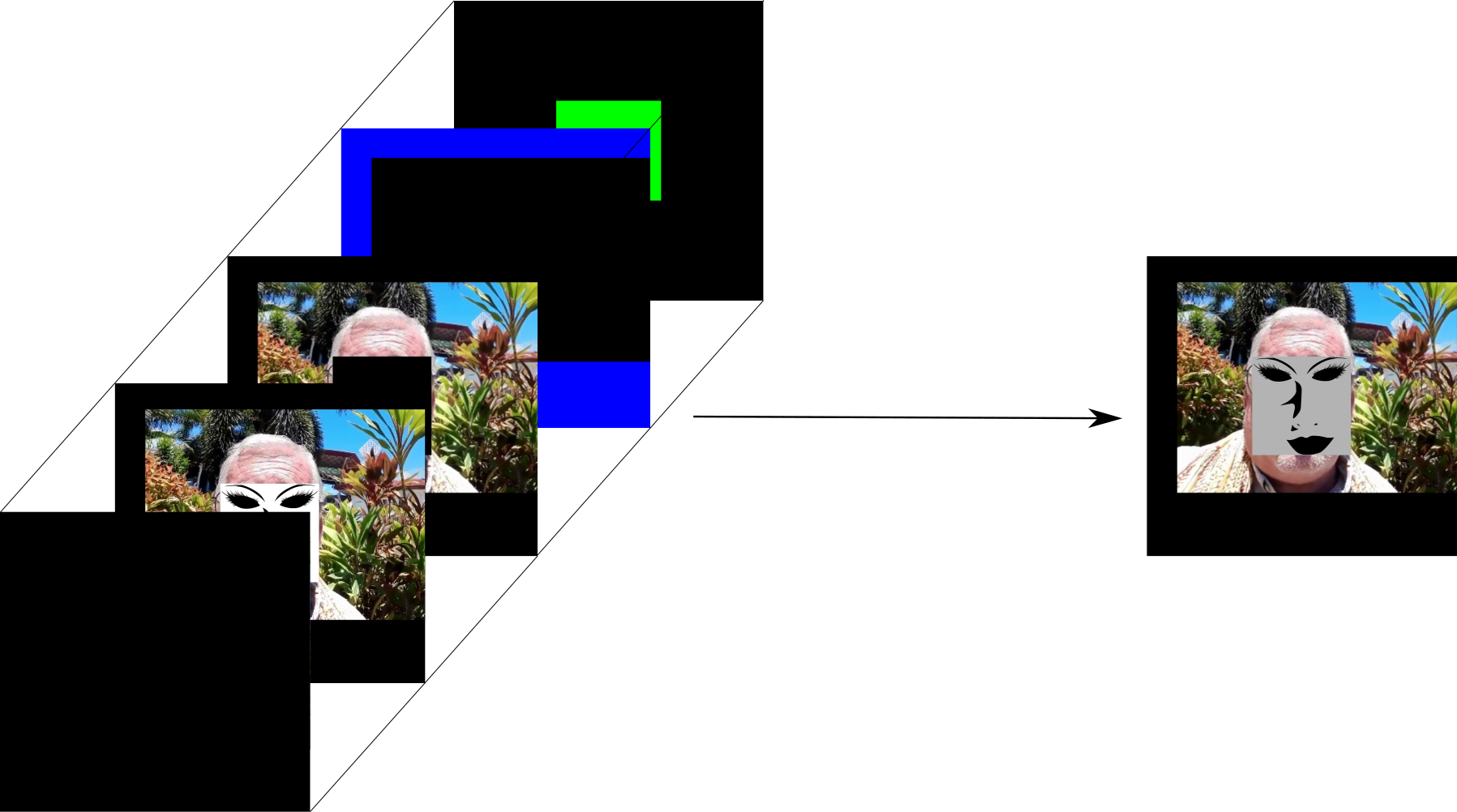}
     \label{fig:burn_in_b}
  \end{minipage}
  }
 \subfloat[]
{
  \begin{minipage}[h]{0.32\linewidth}
    \centering
    \includegraphics[width=0.9\linewidth]{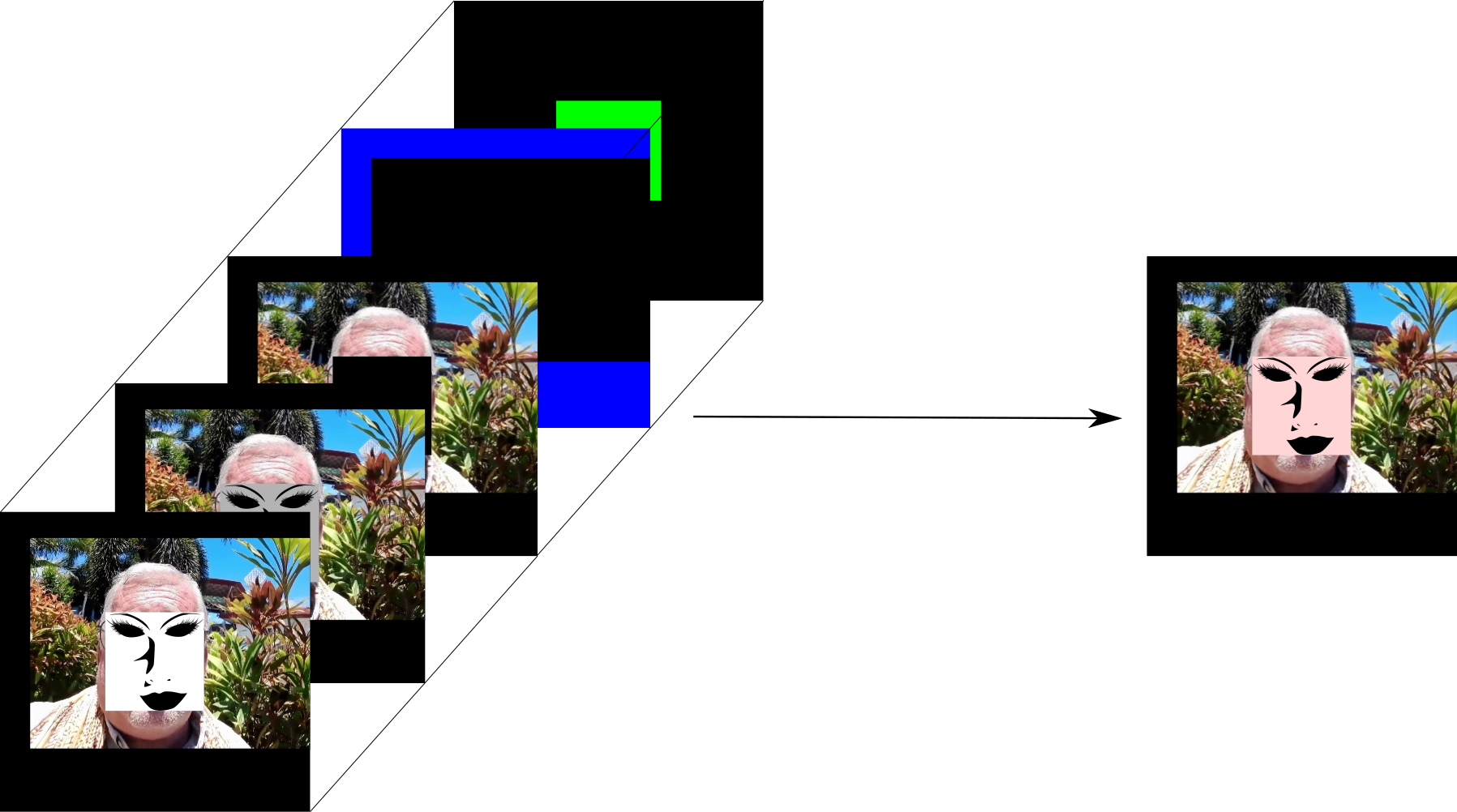}
     \label{fig:burn_in_c}
  \end{minipage}
  }
  \caption{
    Burn--in process:
        \ref{fig:burn_in_a} shows the input tensor for the video model 
        at the first step of the sequence: since there are no past generated images,
        we concatenate tensors filled with zeroes as past images. 
        \ref{fig:burn_in_b} depicts the second time step: we take the image that was
        generated in the first time step and use it as past image.
        \ref{fig:burn_in_c} depicts the third time step: we take the image that was
        generated in the second time step and use this as the second past image.
        The cropped rectangular region is the missing part of the
        face that is to be anonymized by the model, green is the bounding box given by
        the face detector, blue is the border mask.
  }
  \label{fig:burn_in_vis}
\end{figure}

\begin{figure}[tb]
    \centering
    \includegraphics[width=0.9\linewidth]{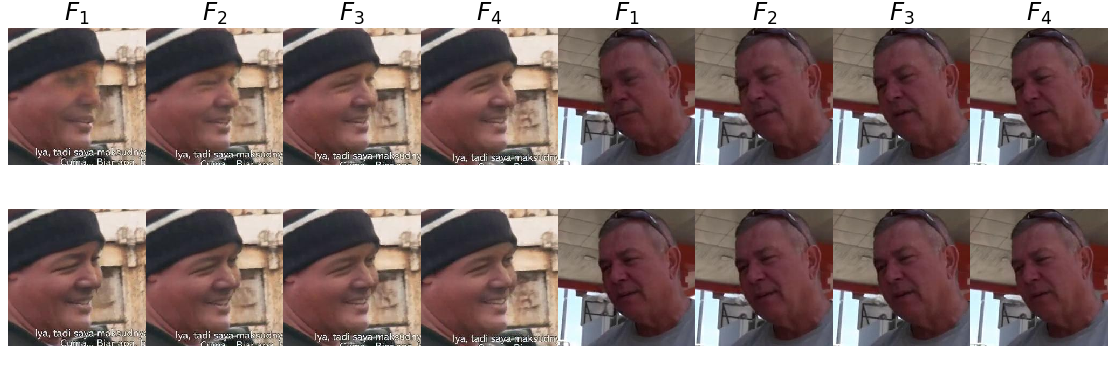}
  \caption{
  The effect of burn--in for inference. Top row: without burn--in, bottom row: 
  with burn--in.
  }
  \label{fig:burn_in_res}
\end{figure}

Fig.~\ref{fig:burn_in_a} depicts the situation when starting to train or infer on a new sequence.
As no previous images are available as an input for the generator when starting with a new sequence, 
we use zero--tensors as substitution during training.
However, when performing inference, there is a noticeable effect of using
zero--tensors at the beginning of a sequence:
since we have no past information to condition the generator on,
the identity of the newly generated faces changes significantly during the first frames.
This can be clearly seen in Fig.~\ref{fig:burn_in_res}.
For this reason, we introduce a stage called \emph{burn--in} for inference: 
Before we generate the first output frame, we create a series of 6 frames, where every subsequent generated frame gets the previous two as inputs. The number of \emph{burn--in} frames is larger than the two conditional inputs, as we need multiple cycles to get rid of the suboptimal contribution of the initial frame.

This can be seen as a kind of bootstrapping since we generate a new face and use 
this newly generated face as artificial past information that we use to condition the next generated face on.
Our experiments in Section~\ref{sec:experiments} reveal that
the burn-in stage not only has a positive effect on subjectively perceived image quality
but also on quantitative evaluation measures, which is why we use it for all inference calculations.

\subsection{Optimization Details}
We use the same losses and regularization as in the image case. Noteworthy is that we had to adjust the weight balance of the different loss terms, due the inclusion of the video discriminator. Also we trained the network with one joint step for discriminator and generator using Adam with a learning rate of 0.0004 and a batch size of 96.

As with the image model,
we monitor a quantitative metric on the validation set to determine when to stop training
and which model to pick for inference. 
As evaluation metric we use the FVD score proposed by \cite{DBLP:conf/nips/vid2vid} and \cite{fvd},
an extension of the FID to videos. The selected model was trained for 1,285,490 mini-batches.

\section{Experimental Evaluation}
\label{sec:experiments}
This section provides an experimental evaluation of both, the image based as well as the video based anonymization networks.

\subsection{Image Models}
\label{image_models}
To verify that the image--based model serves as a valid starting point
for the video model presented in Section~\ref{sec:video_method}
we carry out initial experiments by comparing it to the anonymization GAN presented in \cite{DBLP:conf/isvc/HukkelasML19}.
We train our image model with the FDF dataset released by
\cite{DBLP:conf/isvc/HukkelasML19}. 
We then anonymize the faces of the FDF validation set and calculate the FID score as a quantitative evaluation measure. 
The results can be found in Table~\ref{tab:fid_image_models}.
With our architecture, we achieve state of the art results for facial 
inpainting without using landmarks as an input to the generator.
Given this result we conclude that our model serves as a valid starting point for the video anonymization setting as well. 

\begin{figure}[tb]
\centering
\includegraphics[width=0.99\linewidth]{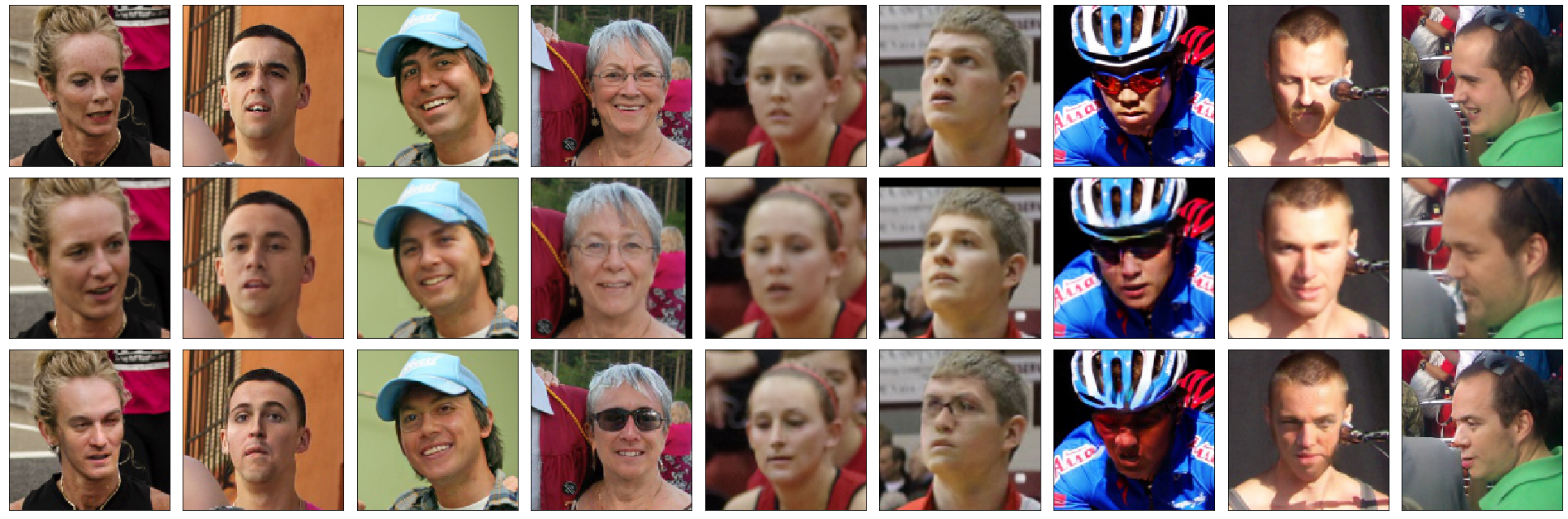}
\caption{Example images generated from the FDF validation set. The first row 
shows the original images, the second row shows
images inpainted by our image model, 
the third shows images inpainted by DeepPrivacy 46M with 
LM \cite{DBLP:conf/isvc/HukkelasML19}.}
\label{fig:img_comparison}
\end{figure}

\begin{table}[tb]
\caption{FID Scores calculated on FDF validation set (lower is better). Our method, trained on the 
FDF training dataset improves the state of the art in landmark--free face inpainting on the 
FDF validation dataset from 3.36 to 1.97.}
\centering
\begin{tabular}{lc}
\toprule
Model & FID score  \\
\midrule
Our Image Model, trained on FDF training dataset \cite{DBLP:conf/isvc/HukkelasML19}  & 1.97 \\
DeepPrivacy 12M with LM \cite{DBLP:conf/isvc/HukkelasML19} & 2.71 \\ 
DeepPrivacy 12M without LM \cite{DBLP:conf/isvc/HukkelasML19} & 3.36 \\ 
DeepPrivacy 46M with LM \cite{DBLP:conf/isvc/HukkelasML19}  & 1.84 \\ 
\bottomrule
\end{tabular}
\label{tab:fid_image_models}
\end{table}

\subsection{Video Models}
In this section, we evaluate the proposed video model in terms of 
quality (via the FVD score \cite{DBLP:conf/nips/vid2vid}, \cite{fvd})
as well as
for temporal coherent face generation across multiple frames for the 
video model.

\subsubsection{The Identity Invariance Score (IdI)}
\label{identity_invariance}
To the best of our knowledge, there is no method measuring the
temporal consistency in a way that humans perceive it.
That is why we also provided videos for the relevant models as supplementary material for visual inspection.
In these videos a lack of temporal consistency can be observed as flickering and warping faces with changing
identities.
To obtain a quantitative metric for temporal coherence,
we postulate that the temporal coherence in generated videos
and an invariance of the identity of the generated faces are correlated, 
i.e., measuring identity invariance
will allow conclusions to be drawn about temporal consistency.
For this reason, we introduce the Identity Invariance (IdI) score.

To measure the average changes of the identity of a generated face in 
a sequence of frames, we proceed in the following way: 
First, we detect the faces with DSFD and calculate the Facenet \cite{Schroff2015} embeddings for each detected face.
This results in sequences of Facenet embeddings with a length of 30 frames per sequence.
We then calculate the squared $\text{L}_2$ distance of the Facenet embeddings between every two subsequent frames.
This provides us a measure to estimate how much the identity of
inpainted faces in two subsequent frames differ, 
which should be in turn correlated with temporal consistency.
We use the squared $\text{L}_2$ distance here because this is what the authors of Facenet use in their paper to determine differences in identities. 
The distances are averaged over the whole sample.

\begin{figure}[bt]
\subfloat[Test Set]
{
  \begin{minipage}[h]{0.32\linewidth}
    \centering
    \includegraphics[width=\linewidth]{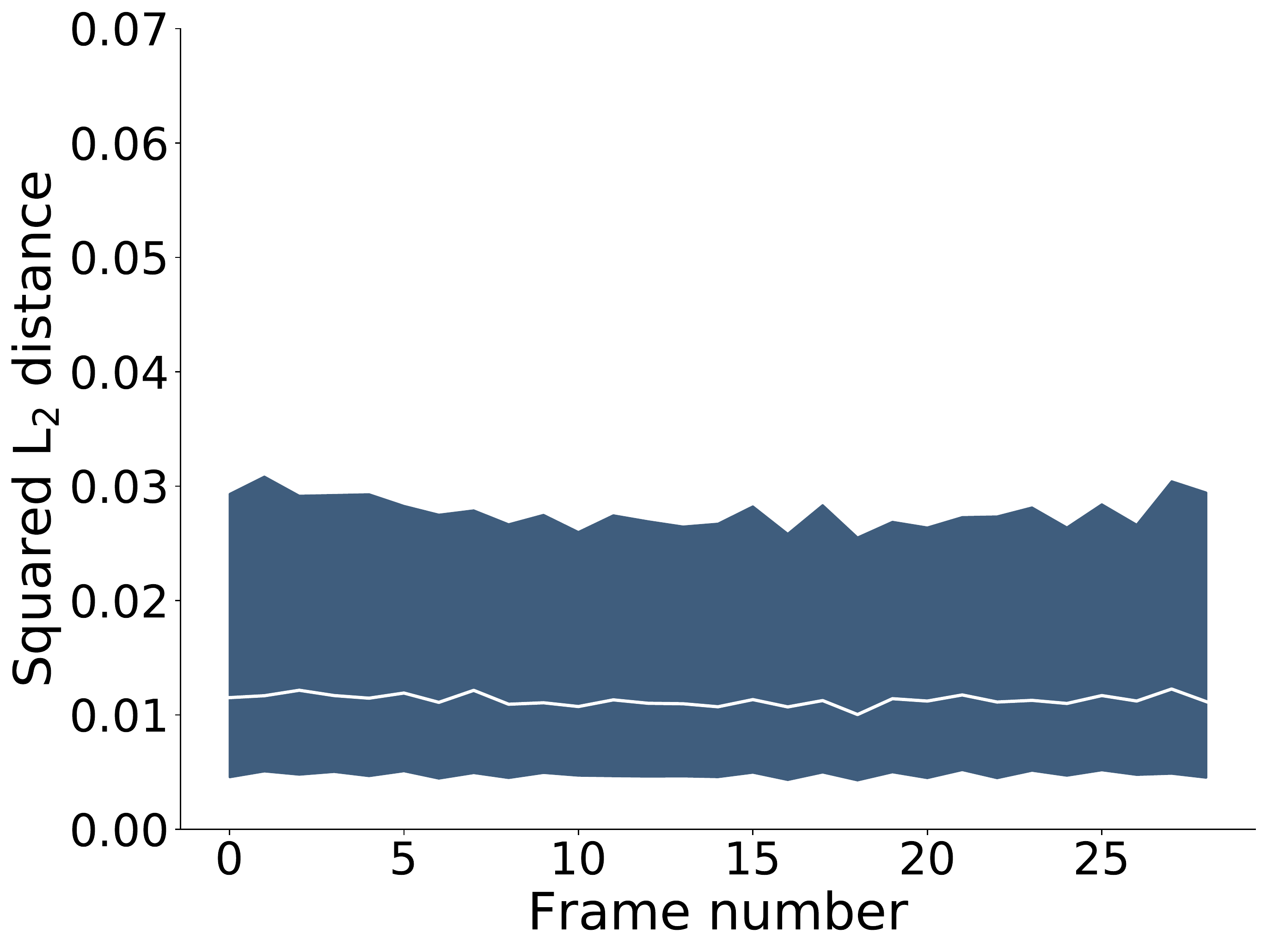}
     \label{fig:resa}
  \end{minipage}
  }
\subfloat[Image Model]
{
  \begin{minipage}[h]{0.32\linewidth}
    \centering
    \includegraphics[width=\linewidth]{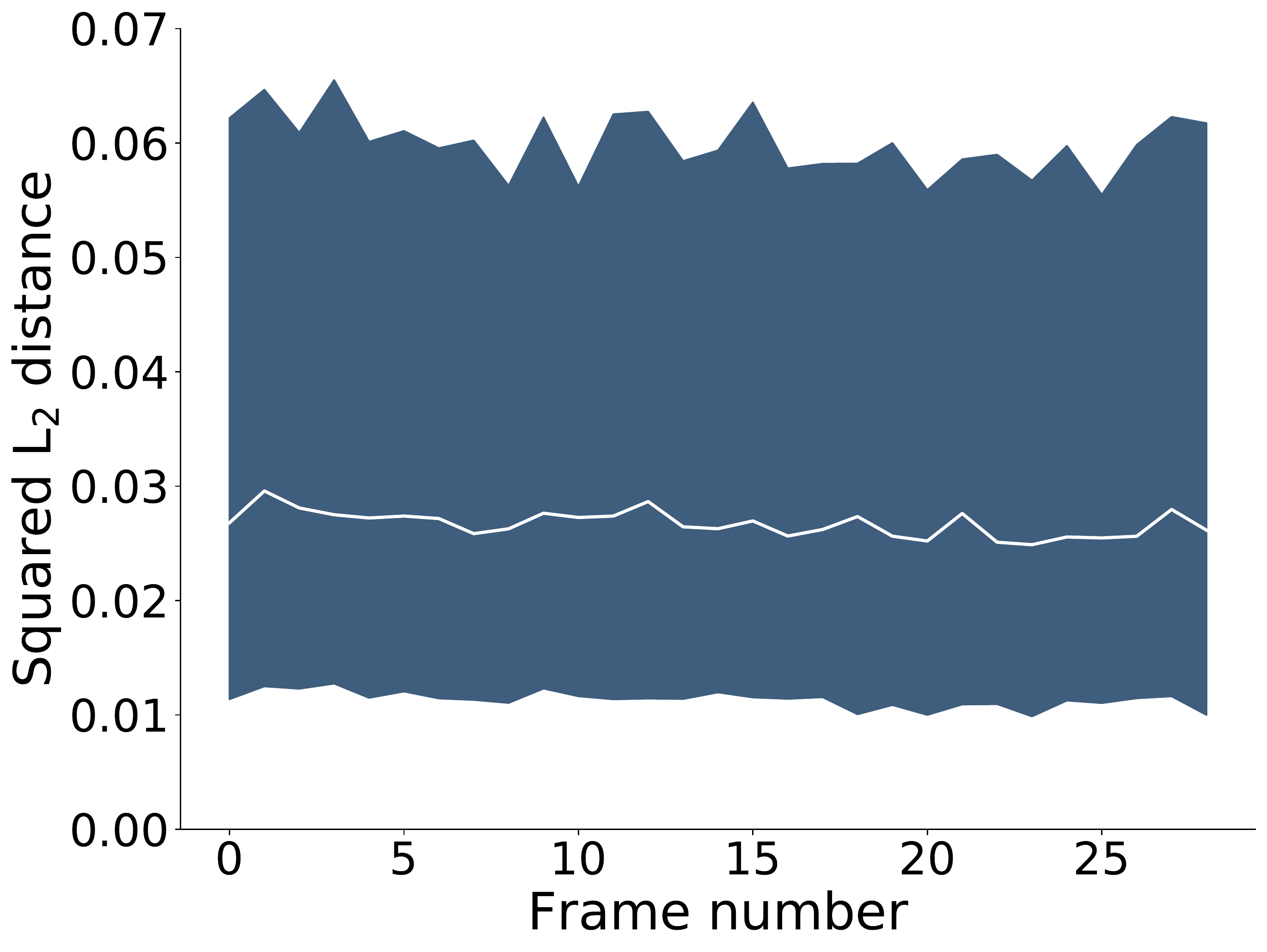}
     \label{fig:resb}
  \end{minipage}
  }
\subfloat[Video Model]
{
  \begin{minipage}[h]{0.32\linewidth}
    \centering
    \includegraphics[width=\linewidth]{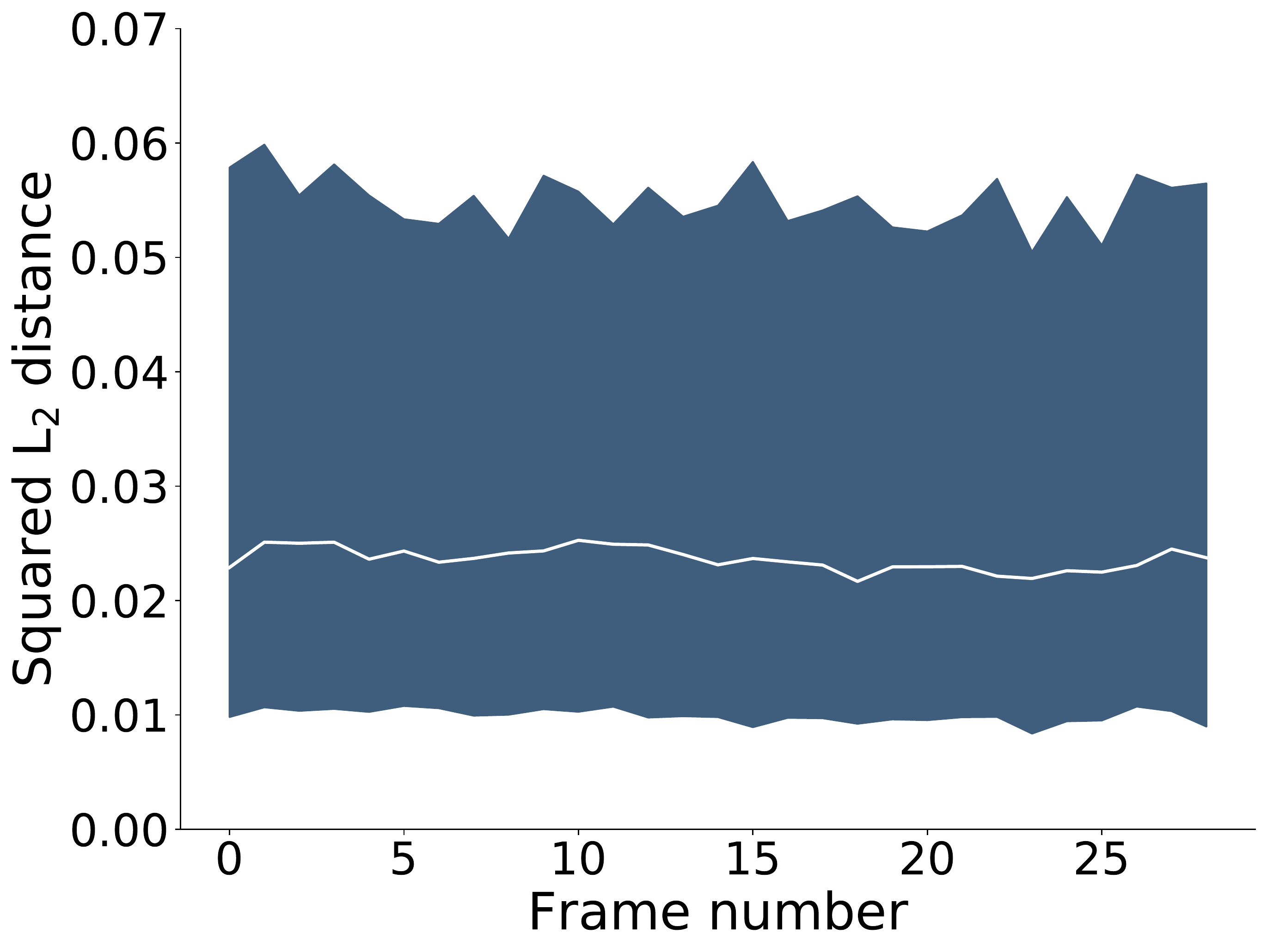}
     \label{fig:resc}
  \end{minipage}
  }
  \caption{These plots show the median along with the IQR of the 
  squared $\text{L}_2$ distance between the Facenet embeddings of 
  2 adjacent frames in a sequence of 30 frames. The distances are 
  aggregated over every sequence in a random sample over the test set,
  which consists of about 50,000 frames. 
  Smaller distances indicate fewer changes of identity
  in the sequence.}
  \label{fig:temporal_coherency}
\end{figure}

To arrive at a point estimate for the temporal consistency 
across the test dataset, 
we take the median of the distances over the whole random sample
for each frame. Plots of these medians over frames can be found in 
Fig.~\ref{fig:temporal_coherency}.
We propose the \emph{IdI Score} as the ratio of the median of squared $\text{L}_2$ distances
of real frames over 
the median of squared $\text{L}_2$ distances of generated frames.
The reason why we take the ratio over the distances of the original 
test set is that there are naturally occuring changes in the Facenet
embeddings when a face moves. To take this into consideration, 
we normalize by the distance of the real distances. 

Table \ref{tab:fvd_video_models} shows the IdI scores of our
image and video datasets and for the real faces in the test set. 

\begin{table}[bt]
\caption{Metrics
calculated on our test set (for the FVD scores and the distances,
lower is better, 
for the IdI score, higher is better)}
\centering
\begin{tabular}{lccc}
\toprule
Model & FVD score & Med. sq. $\text{L}_2$ & IdI score \\
\midrule
Test Set & n/a & 0.0113 & 1.00\\
Our Image Model & 110 & 0.0268 & 0.42  \\
Our Video Model & {\bf 59} & 0.0237 & {\bf 0.48}\\
\bottomrule
\end{tabular}
\label{tab:fvd_video_models}
\end{table}

\section{Conclusion}
\label{sec:conclusion}
We propose and release a large scale facial video dataset diverse in terms of age, gender, ethnicity and head poses
for training and testing video based anonymization methods.
Given this dataset, we develop a GAN-based method for inpainting generated faces 
into anonymized images and videos that is not dependent on any landmarks. 
On single frame images, we improved the state of the art FID score from 3.36 to 1.97 for landmark--free face inpainting.
When working with video data our approach is able to generate coherent faces across sequences. It reduces identity shift across individual frames in comparison to an image (single frame) based inpainting model.
In order to measure the degree of coherency, we introduced the identity invariance score and show in experimental evaluations that our video model helps to keep generated faces consistent over a sequence of frames.

We hope that the provided dataset as well as our baseline models help to foster further research in the field of image preserving face anonymization.


{\small
\bibliographystyle{ieee}
\bibliography{sample_FG2021}
}

\end{document}